%% file: main.tex
\documentclass[10pt,twocolumn,letterpaper]{article}

\usepackage[pagenumbers]{iccv} % To force page numbers, e.g. for an arXiv version
\usepackage{booktabs} % For professional table rulings
\usepackage{siunitx} % For aligned numbers
\usepackage{xcolor} % For row colors
\usepackage{colortbl} % For alternating row colors
\usepackage{tabularx}
\usepackage{algorithm}
\usepackage{algpseudocode}

\definecolor{iccvblue}{rgb}{0.21,0.49,0.74}
\usepackage[pagebackref,breaklinks,colorlinks,allcolors=iccvblue]{hyperref}

\def\paperID{12587} % *** Enter the Paper ID here
\def\confName{ICCV}
\def\confYear{2025}

\title{Mind the Gap: Benchmarking Spatial Reasoning in Vision-Language Models}%SRBench: A Comprehensive Benchmark to Bridge the Gap between Machine and Human Spatial Reasoning.

\author{Ilias Stogiannidis, Steven McDonagh, Sotirios A. Tsaftaris\\
The University of Edinburgh\\
Edinburgh, UK\\
{\tt\small \{i.stogiannidis, s.mcdonagh, s.tsaftaris\}@ed.ac.uk}
}

\begin{document}
\maketitle
\input{sec/0_abstract}

\input{sec/1_intro}
\input{sec/2_spatial_reasoning}

\input{sec/3_framework}
\input{sec/4_eval}

\input{sec/5_related_work}
\input{sec/6_conclusion}
{
    \small
    \bibliographystyle{ieeenat_fullname}
    \bibliography{main}
}

\input{sec/X_suppl}

\end{document}

%% file: sec/0_abstract.tex
\begin{abstract}

Vision-Language Models (VLMs) have recently emerged as powerful tools, excelling in tasks that integrate visual and textual comprehension, such as image captioning, visual question answering, and image-text retrieval. However, existing benchmarks for VLMs include spatial components, which often fail to isolate spatial reasoning from related tasks such as object detection or semantic comprehension. In this paper, we address these deficiencies with a multi-faceted approach towards understanding spatial reasoning. Informed by the diverse and multi-dimensional nature of human spatial reasoning abilities, we present a detailed analysis that first delineates the core elements of spatial reasoning: spatial relations, orientation and navigation, mental rotation, and spatial visualization, and then assesses the performance of these models in both synthetic and real-world images, bridging controlled and naturalistic contexts. We analyze 13 state-of-the-art Vision-Language Models, uncovering pivotal insights into their spatial reasoning performance. Our results reveal profound shortcomings in current VLMs, with average accuracy across the 13 models approximating random chance, highlighting spatial reasoning as a persistent obstacle. This work not only exposes the pressing need to advance spatial reasoning within VLMs but also establishes a solid platform for future exploration. Code available on \href{https://github.com/stogiannidis/srbench}{GitHub}\footnote{\url{https://github.com/stogiannidis/srbench}} and dataset available on \href{https://huggingface.co/datasets/stogiannidis/srbench}{HuggingFace}.\footnote{\url{https://huggingface.co/datasets/stogiannidis/srbench}}

\end{abstract}

%% file: sec/1_intro.tex
\section{Introduction}
\label{sec:intro}

Vision-Language Models (VLMs) have recently emerged as powerful tools, excelling in tasks that integrate visual and textual comprehension such as image captioning, visual question answering, and image-text retrieval~\cite{liu2023visual, dubey2024llama}. Developed through extensive pretraining on diverse datasets, these models have demonstrated a remarkable ability to interpret complex interactions between visual information and language~\cite{radford2021learning}. However, despite these advances, a critical capability remains largely unaddressed: spatial reasoning.
\begin{figure}[!t]
    \centering
    \includegraphics[width=\columnwidth]{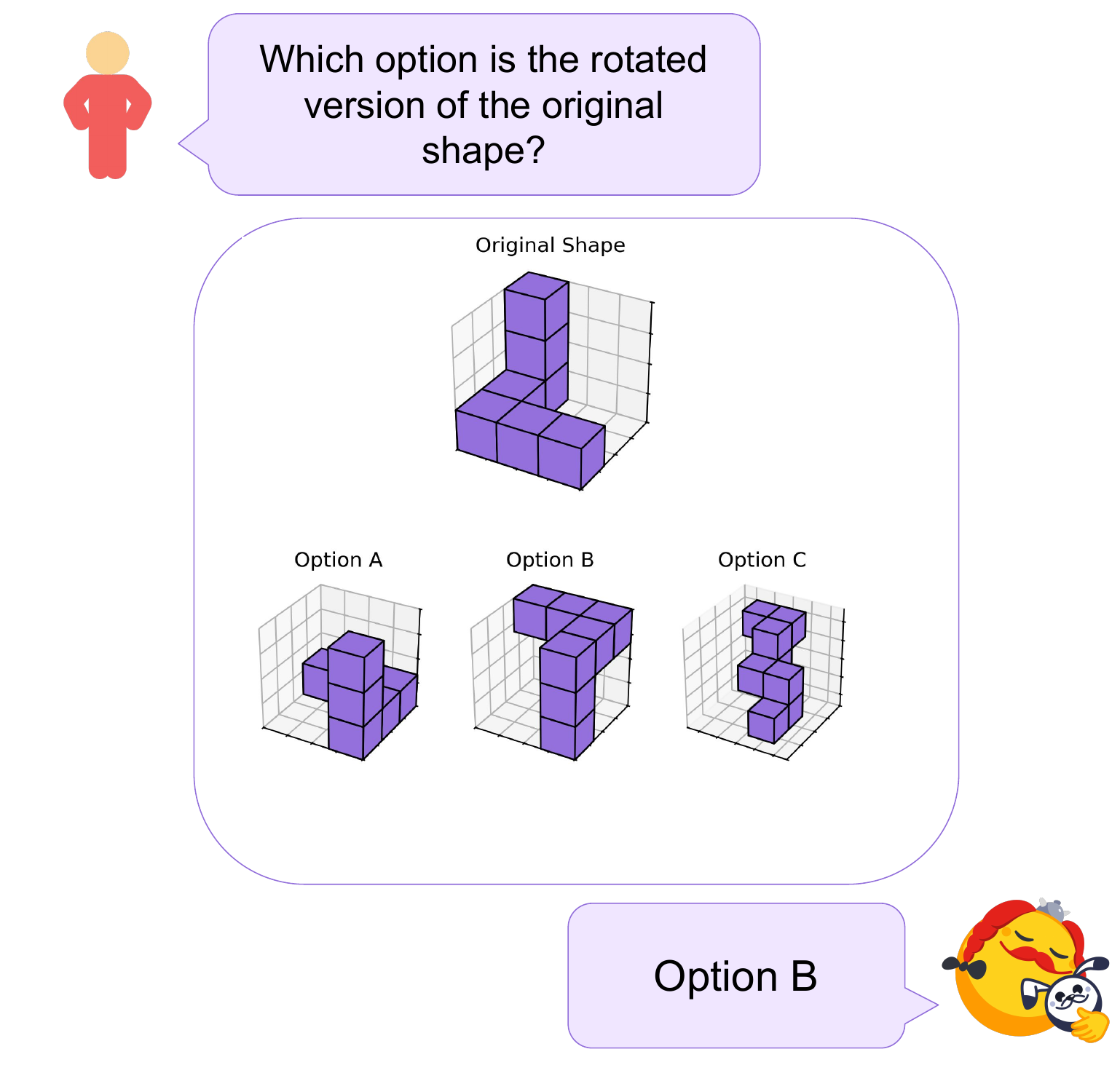}
    \caption{\textbf{Example of VLM Responses to Mental Rotation Tasks:} This example underscores a notable constraint of existing VLMs, which have difficulty in precisely understanding potential rotations of the objects shown, exposing a considerable deficiency in their spatial reasoning skills.}
    \label{fig:intro}
    \vspace{-5pt}
\end{figure}

Spatial reasoning involves comprehending and analyzing the locations, orientations, and relations of objects within a scene—skills that are natural to humans but pose significant challenges for deep learning models~\cite{shiri-etal-2024-empirical, chen2024spatialvlm, cheng_spatialrgpt_2024}. By fostering strong spatial intelligence, deep learning models can improve their ability to make intricate decisions that necessitate an understanding of geometric transformations and spatial context. 
% Although current benchmarks for VLMs include spatial components, they often confuse spatial reasoning with related tasks like object detection or semantic interpretation.
While current benchmarks for VLMs incorporate tasks related to spatial reasoning, they often prioritize elements such as object detection or semantic interpretation~\cite{liu2024mmbenchmultimodalmodelallaround}, leaving the core challenges of spatial cognition underexplored. Additionally, many datasets focus exclusively on specific aspects of spatial reasoning, sometimes augmenting visual data with extra elements such as point clouds, scene graphs, or depth maps to deepen spatial context~\cite{chen2024spatialvlm, cheng2025spatialrgpt}. The absence of a precise definition and structured evaluation approach for spatial reasoning in VLMs has hindered the creation of models capable of attaining human-level spatial reasoning.
Humans naturally excel at spatial reasoning, a cognitive faculty that manifests  in development and underpins our ability to engage with and navigate intricate environments~\cite{johnson1987body, newcombe2000making}. This competence includes a variety of skills, from calculating distances and orienting things in three dimensions to combining various sensory inputs in a smooth manner.  It has been established via rigorous empirical studies in cognitive science and neuroscience that our brain architecture is suited to process spatial relationships.  These findings constantly confirm that spatial reasoning is deeply ingrained in our perceptual and motor systems and is not just essential for daily tasks~\cite{burgess2008spatial, husain2007space}.  By attempting to understand and reproduce these natural human abilities, we can shed light on the shortcomings of current VLMs and drive the development of computer systems that exhibit closer to human-like spatial awareness.
% This proficiency encompasses a spectrum of abilities, ranging from the estimation of distances and the orientation of objects within three-dimensional space to the seamless integration of multiple sensory cues. Our cerebral architecture is calibrated to process spatial relationships, a capability that has been substantiated through rigorous empirical investigations in cognitive science and neuroscience. These studies consistently affirm that spatial reasoning is not merely indispensable for routine activities but is profoundly embedded within the fabric of our perceptual and motor systems~\cite{burgess2008spatial, husain2007space}. By striving to comprehend and replicate these innate human competencies, we are well-positioned to illuminate the deficiencies inherent in contemporary vision-language models (VLMs) and to propel the advancement of computational systems that more closely approximate human-like spatial acumen.

Performant spatial reasoning is crucial beyond lab tests, enabling computational agents in fields such as robotics, autonomous navigation, and augmented reality to adapt to complex real-world environments~\cite{venkatesh2021spatial}. It enhances VLMs for interpreting visual scenes anthropomorphically and drives advancements in technologies requiring precise spatial understanding.
% Furthermore, the significance of spatial reasoning extends far beyond the controlled settings of laboratory-based benchmarks. In applied contexts such as robotics, autonomous navigation systems, and augmented reality interfaces, a sophisticated grasp of spatial dynamics empowers computational agents to interact intelligently with, and adapt to, the multifaceted complexities of dynamic real-world environments~\cite{venkatesh2021spatial}. The enhancement of spatial reasoning capacities within vision-language models not only facilitates a more anthropomorphic interpretation of visual scenes but also serves as a catalyst for transformative innovations in technological domains that are contingent upon precise spatial cognizance. This broader applicability underscores the imperative of advancing our understanding of spatial reasoning beyond theoretical constructs, anchoring it firmly within practical, operational frameworks.
The present study aims to fill these gaps by a thorough and systematic analysis of spatial reasoning in the context of VLMs. We provide a thorough analysis that begins with the characterization of the basic components of spatial reasoning, namely spatial relations, orientation and navigation, mental rotation, and spatial visualization. After laying this conceptual foundation, we assess the models' performance on a wide range of visual stimuli, including both artificially generated images and images from naturalistic environments. By performing a discrete evaluation of each component, we gain a more nuanced understanding of the unique advantages and disadvantages that exist in current VLMs and help us understand their operational capabilities.
In summary, our contributions are as follows:
\begin{itemize}
    \item \textbf{Precise Definition of Spatial Reasoning and its components:} %We define spatial reasoning for VLMs with key components: spatial relations (e.g., identifying if one object is left of another), orientation and navigation (e.g., determining an object's facing direction), mental rotation (e.g., recognizing rotated object versions), and spatial visualization (e.g., predicting an object's transformation after actions).   
    We formalize spatial reasoning in VLMs through key components: spatial relations (e.g., object positioning), orientation and navigation (e.g., directionality), mental rotation (e.g., invariant object recognition), and spatial visualization (e.g., transformation prediction).
    \item \textbf{Extensive Benchmark:} 
    %We introduce an innovative benchmark for spatial reasoning, integrating both programmatically and GenAI-generated synthetic images with real-world scenes, bridging the gap between controlled experiments and natural settings.
    We present a novel spatial reasoning benchmark, combining programmatically generated, GenAI-synthesized, and real-world images to bridge controlled evaluation and real-world applicability.
    \item \textbf{Component-Specific Evaluation:} %Our benchmark uniquely assesses individual spatial reasoning components, providing specific insights into VLMs' strengths and weaknesses.
    Our benchmark uniquely isolates spatial reasoning components, offering precise insights into VLMs' strengths and limitations.
    \item \textbf{Extensive Model Evaluation:} We assess 13 advanced VLMs, discovering that their typical spatial reasoning performance hovers around random chance, pointing out ongoing challenges. %These models exhibit minor enhancements when using images similar to their training data, implying the model
\end{itemize}

%% file: sec/2_spatial_reasoning.tex
\begin{figure*}[!t]
    \centering
    \includegraphics[width=0.8\textwidth]{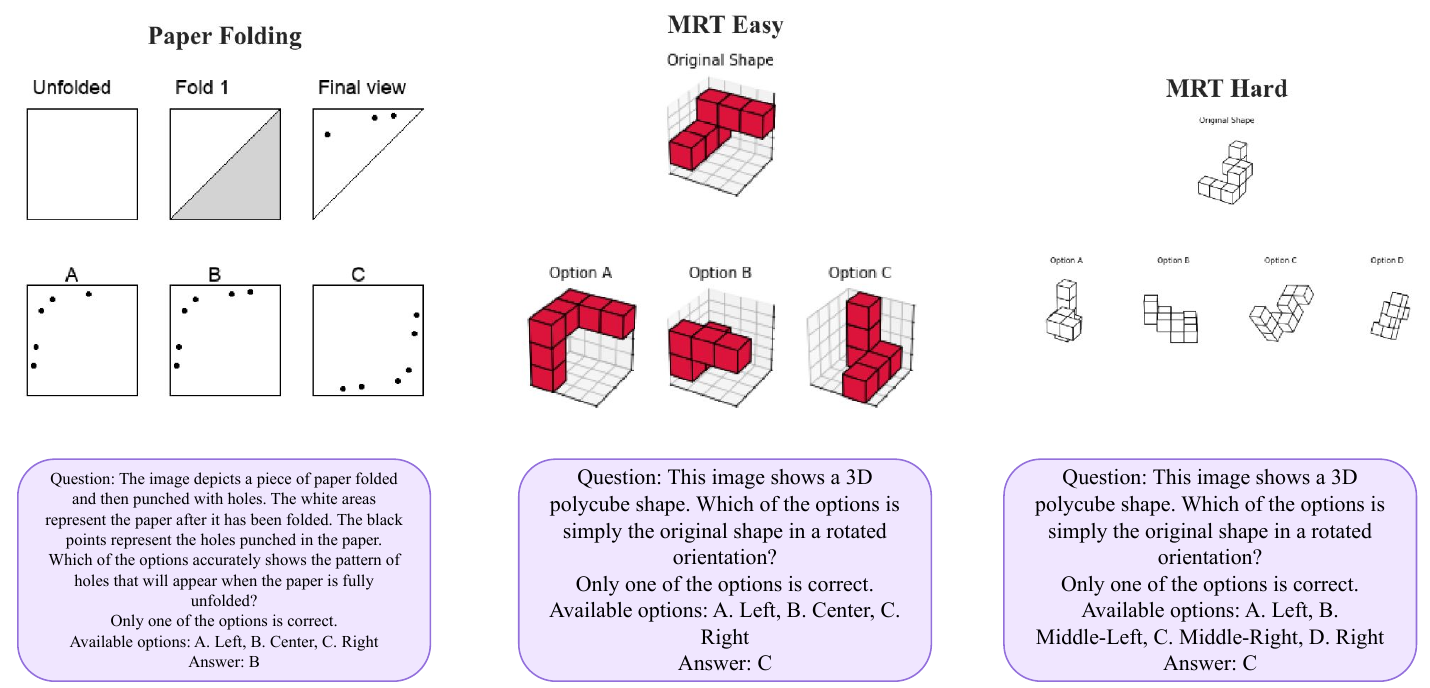}
    \caption{Images from our benchmark created algorithmically, drawing inspiration from cognitive tests. Left image depicts paper folding, the middle shows the easy MRT version, and the right displays the hard variant.}
    \label{fig:bench_s}
    \vspace{-10pt}
\end{figure*}
\section{Spatial Reasoning: Definition and Components}

Spatial reasoning is a fundamental cognitive ability concerned with understanding and manipulating spatial relationships between objects and oneself within a space \cite{linn1985emergence}. It encompasses the mental skills involved in visualizing, transforming, and reasoning about spatial information. This ability is not monolithic but rather comprises several distinct yet interrelated components \cite{montello2005cognitive}. In computer science, spatial reasoning is increasingly relevant to fields such as robotics, computer vision, and geographic information systems, where algorithms and systems must effectively process and interpret spatial data \cite{stock2008spatial}. Understanding these components is crucial for designing effective computational models of spatial intelligence.

\subsection{Mental Rotation}

Mental rotation is a specific and well-studied aspect of spatial visualization, which refers to the ability to mentally rotate two-dimensional (2D) and three-dimensional (3D) objects \cite{shepard1971mental}.  It involves imagining an object as it would appear if rotated in space, and is often measured by tasks requiring individuals to determine if two objects are the same, but presented at different orientations~\cite{vandenberg1975mental}. \citet{pellegrino1998cognitive} showed that mental rotation is a distinct spatial ability, separate from other spatial skills such as spatial perception. Mental rotation tasks are widely used in cognitive psychology to study spatial processing and have implications for understanding sex-related differences in spatial abilities, as well as neural substrates of spatial cognition~\cite{wraga2000mental}. While closely related to spatial visualization, mental rotation is often considered a more constrained and specific type of spatial transformation, focusing primarily on rotational transformations.

\subsection{Spatial Visualization}

Spatial visualization goes beyond simple perception and involves the capacity to mentally manipulate and transform spatial information~\cite{mcgee1979human}. It is often defined as the ability to process complex spatial information and imagine spatial transformations, such as mentally rotating objects, folding shapes, or understanding movements through space \cite{ekstrom1976manual}.  Spatial visualization is considered a dynamic spatial skill, requiring the transformation of visual-spatial representations to derive new spatial configurations \cite{logan1996spatial}. This component is particularly important in fields such as engineering and design, where professionals must mentally simulate the assembly or operation of complex systems. In computer graphics and virtual reality, spatial visualization skills are essential for creating immersive and interactive 3D environments, allowing users to navigate and manipulate virtual objects in a spatially meaningful way \cite{bowman20043d}.

\subsection{Spatial Orientation and Navigation}

Spatial orientation is a complex cognitive function that enables us to navigate intricate surroundings by actively integrating various sensory inputs, such as visual, vestibular, proprioceptive, and auditory signals, to construct and continually refine internal spatial maps. Studies indicate that these maps can be divided into two main reference frames: \emph{egocentric}, which records spatial data in relation to one's own body (e.g., ``the book is on my left''), and \emph{allocentric}, which organizes spatial data based on the relationships among landmarks in the environment, regardless of the observer's position \cite{darken1999spatial, Wang2002-lh}. This duality facilitates flexible navigation; path integration allows individuals to compute their position by updating self-motion cues, and landmark-based encoding helps create a more stable, map-like allocentric space representation \cite{klatzky1998spatial}. In robotics, these principles guide autonomous systems like Simultaneous Localization and Mapping (SLAM), where data from cameras, LiDAR, and inertial units are combined using probabilistic models for accurate real-time mapping and self-localization, even in GPS-denied or dynamic environments \cite{thrun2005probabilistic}. These interdisciplinary insights show that both egocentric and allocentric frameworks are crucial for human spatial cognition and adaptation, as well as for designing artificial systems interacting with complex environments.

\subsection{Spatial Relations}

Spatial relations encompass the ability to understand and reason about relationships between multiple objects within a space \cite{newcombe2000developing}. This involves processing different types of spatial relations, including topological relations (for example, adjacency, containment, connectivity), projective relations (for example, above / below, left / right, front / behind), and metric relations (e.g., distance, size, volume) \cite{freksa2013spatial}. Reasoning about spatial relationships is fundamental for tasks such as spatial planning, solving spatial puzzles, and understanding spatial analogies. In geographic information systems (GIS), the ability to computationally represent and reason with spatial relations is paramount for spatial queries, spatial analysis, and automated map interpretation~\cite{bennett1998qualitative}.  The formalization of spatial relations in computational terms is an ongoing area of research in artificial intelligence.

% \begin{figure}[!thb]
%     \centering
%     \includegraphics[width=1\linewidth]{ICCV2025-Author-Kit//img/mrt_ex.pdf}
%     \caption{\textbf{Illustration of the MRT Hard Task:} The top row depicts a polycube shape at its original version, while the bottom row presents four possible correct outcomes after rotation.}
%     \label{fig:mrt}
% \end{figure}

Understanding these distinct yet interconnected components provides a more nuanced framework for both cognitive and computational investigations of spatial reasoning. 

% Future research should continue to explore the interaction between these components and their respective roles in complex spatial tasks, paving the way for more sophisticated and human-like spatial reasoning capabilities in artificial intelligence systems.

%% file: sec/3_framework.tex
\section{Data Creation}

In contrast to previous benchmarks which test a few aspects of spatial reasoning, our benchmark draws inspiration from human cognition studies~\cite{HEGARTY2010265, darken1999spatial, Wang2002-lh} and is designed to evaluate VLMs under a comprehensive set of the basic principles of human spatial reasoning.

\subsection{Mental Rotation}

To assess a VLM's capability to mentally rotate an object, we draw upon the mental rotation test~\cite{cooper1975mental}, a test designed to measure this ability in humans. The Mental Rotation Test (MRT) involves a participant comparing two 3D objects (or letters), often rotated along some axis, to determine if they are the same or mirror images~\cite{Shepard1971-np}. Typically, pairs of images in the test are rotated by specific angles (e.g. 0\textdegree, 60\textdegree, 120\textdegree, or 180\textdegree). A fixed number of these pairs will feature identical images simply rotated, while others will be mirrored versions. Participants are evaluated by the researcher based on their speed and precision in differentiating between mirrored and non-mirrored pairs~\cite{Caissie2009-xp}. We proceed by creating similar images to those found in MRT~\cite{cooper1975mental}, manually crafting five polycube shapes. %as depicted in Figure~\ref{fig:bench_s}. 
Subsequently, we construct an image displaying the original shape in the top row, with four potential mirrored versions of that shape in the bottom row. Among these four options, one is the same shape rotated by either 0, 60, 90, or 120 degrees. The other three include two mirrored forms of the shape, each rotated, and one randomly selected, differently rotated object. The shapes are white and devoid of any background, forming what we term the \textit{mrt-hard} subset. An example from \textit{mrt-hard} is depicted in Figure~\ref{fig:bench_s} (right). Considering that these images might not offer sufficient visual cues to the model, we develop another test version with colored shapes and a 3D Cartesian grid as the background. Furthermore, we decrease the number of candidates to three by eliminating one mirrored version of the shape. This variant is referred to as the \textit{mrt-easy} subset. We generate 200 images from both subsets. Figure~\ref{fig:bench_s} (middle) shows an example of \textit{mrt-easy} images.

\begin{figure*}[!t]
  \centering
  \includegraphics[width=0.8\textwidth]{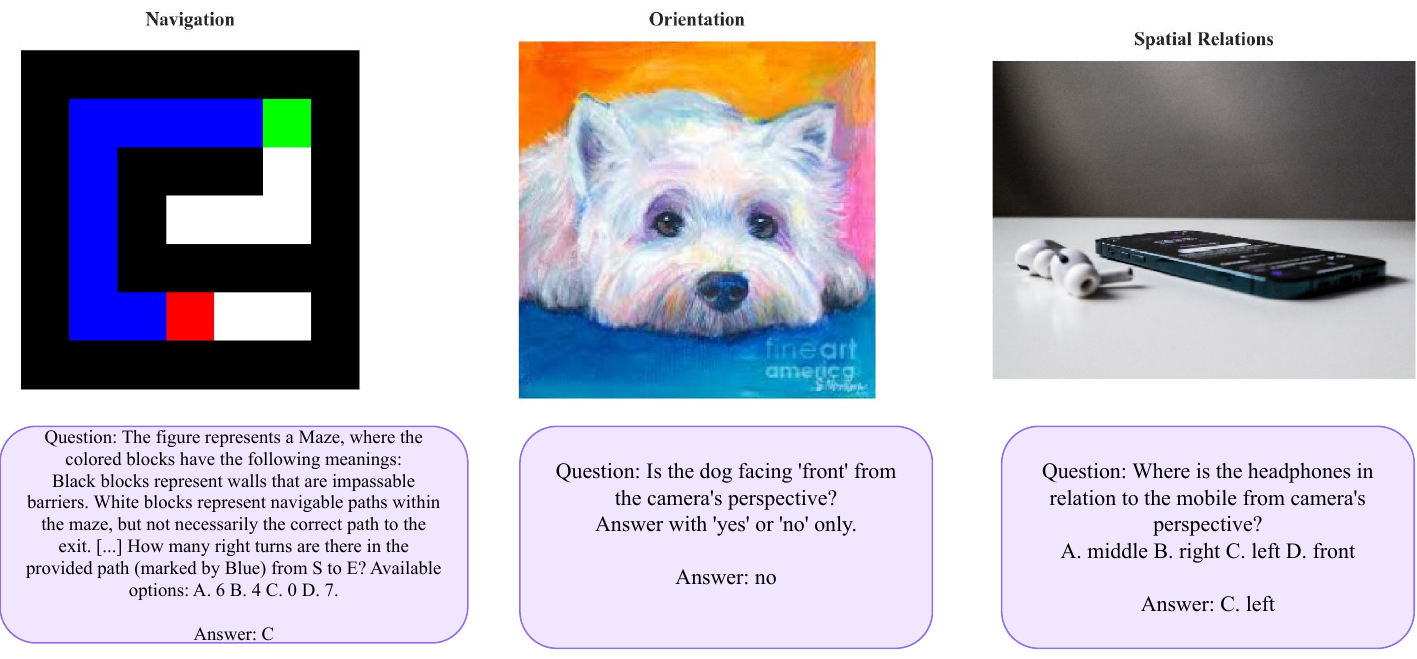} 
  \vspace{-3pt}
  \caption{Examples sampled from other benchmarks are depicted as follows: Left is \emph{Maze-Nav}~\cite{wang2025picture}, the center displays \emph{EgoOrientBench}~\cite{jung2024right}, and on the right is \emph{Spatial-Obj}~\cite{shiri-etal-2024-empirical}.}
  % \vspace{-12pt}
  \label{fig:bench}
\end{figure*}

\subsection{Spatial Visualization}

The Paper Folding Test is a non-verbal reasoning assessment commonly used in psychometric evaluations. It assesses spatial visualization abilities, requiring individuals to mentally manipulate folded paper and identify the location of holes punched through it after unfolding~\cite{ekstrom1976manual}.  This test is considered a measure of spatial orientation and visualization, distinct from spatial perception~\cite{mcgee1979human}~.  Performance on the Paper Folding Test is often correlated with success in fields requiring strong spatial reasoning skills, such as engineering, architecture, and design~\cite{carroll1993human}. The format of the test typically involves a series of multiple-choice questions, each presenting a sequence of paper folds and hole punches, followed by several unfolded paper options, only one of which is correct. We replicate the paper-folding test by constructing an image where the top row displays a square representing a piece of paper being folded. To the right of this image, the folding process is visually depicted by redrawing the entire square and adding a line to indicate where the fold occurs, such as in the center, while shading the folded portion in light gray. Following this depiction, we puncture holes in the final folded view, marking them as black dots on the paper. Consequently, in the bottom row, three different unfolded paper images are shown, illustrating the unfolded paper with final hole positions. Among them, one is correct, another lacks a hole, and the last is accurately mirrored. We can produce an arbitrary number of folds and holes, but for simplicity, we restrict ourselves to experimenting with one or two folds and one to three holes. Moreover, we employ straightforward folding techniques, recursively folding the paper either vertically, horizontally, or diagonally at the center. Figure~\ref{fig:bench_s} (left) depicts an example of these images.

\subsection{Spatial Relations}
To assess VLMs' capability in recognizing spatial relationships, we employ a sample of the \emph{Spatial-Obj} dataset~\cite{shiri-etal-2024-empirical}. This benchmark comprises 2,000 multiple-choice queries aimed at testing how effectively VLMs interpret spatial interactions between objects within images. Developed via a dual-phase annotation method, the dataset includes natural image-based inquiries concerning the spatial interplay of one or two objects. During the initial annotation phase, three annotators generated question-answer sets using templates, followed by a review and correction process by ten evaluators in the subsequent phase. Spatial-Obj encompasses 36 typical spatial relations (e.g., ``right of'', ``left of'', ``attached to'', ``touching''). GPT-4o has been used to organize samples into visual categories such as object localization, orientation and direction, viewpoints, as well as positional/relational context, all presenting considerable challenges for VLMs.

\subsection{Orientation and Navigation}

Spatial orientation plays a crucial role in navigation-related activities. We assess this navigating ability using the \emph{Maze-Nav} component of SpatialEval~\cite{wang2025picture}. This dataset is crafted to test navigation capabilities within intricate environments resembling mazes, represented by colored blocks: green signifies the starting point (S), red marks the exit (E), black indicates walls that cannot be passed, white denotes walkable paths, and blue highlights the correct path from S to E. It is available in both textual (ASCII) and visual formats, with the goal being to navigate from S to E by following the blue path while using only up, down, left, and right directions. Tasks include answering queries such as counting the number of turns or identifying the spatial relationship between S and E, which are straightforward for humans but challenging for current VLMs. 
For evaluating orientation comprehension, we examine 400 question-and-answer pairs from \emph{EgoOrientBench}~\cite{jung2024right}. The annotation method tackles inconsistencies in object orientation labeling within VLMs by creating a unified egocentric system. Utilizing current developments in embodied AI that emphasize user-centric viewpoints, the authors formulated an eight-class orientation taxonomy (Front, Back, Left, Right, Front-Left, Front-Right, Back-Left, and Back-Right) that consistently positions objects with respect to the user (or camera) perspective. In addition, to facilitate easier evaluation, we sample only those questions whose answers are limited to ``yes'' or ``no''. This egocentric alignment not only addresses the problem of unclear orientation annotations that hinder VLMs' spatial understanding but also bolsters their use in real settings where user perspective is key. The approach is in line with the growing trend of egocentric datasets in AI, establishing a structured system that enhances orientation interpretation consistency in various applications.
% For evaluating orientation comprehension, we examine 400 question-and-answer pairs from EgoOrientBench~\cite{jung2024right}. The annotation method tackles inconsistencies in object orientation labeling within VLMs by creating a unified egocentric system. Utilizing current developments in embodied AI that emphasize user-centric viewpoints, the authors formulated an eight-class orientation taxonomy (Front, Back, Left, Right, Front-Left, Front-Right, Back-Left, and Back-Right). This taxonomy consistently positions objects concerning the user's or camera's perspective. This egocentric alignment not only addresses the problem of unclear orientation annotations that hinder VLMs' spatial understanding but also bolsters their use in real settings where user perspective is key. The approach is in line with the growing trend of egocentric datasets in AI, establishing a structured system that enhances orientation interpretation consistency in various applications.
% \begin{figure}[t]
%     \centering
% \includegraphics[width=\columnwidth]{ICCV2025-Author-Kit//img/aspect_distribution.pdf}
%     \vspace*{-7pt}
% \caption{\textbf{Benchmark Component Distribution:} This chart presents the statistical breakdown of various spatial cognition aspects within our benchmark, including Spatial Relations, Navigation, and Orientation (each at 22.2\%), alongside Mental Rotation (Easy \& Hard) and Paper Folding (each at 11.1\%).}
%     \label{fig:distribution}    
% \end{figure}

%% file: sec/4_eval.tex
\begin{table*}[!htbp]
\centering
\small
\renewcommand{\arraystretch}{1.0}
\begin{tabularx}{\textwidth}{p{3.1cm} *{7}{>{\centering\arraybackslash}X}}
\toprule
\rowcolor{violet!25}
\textbf{Model} & \textbf{Paper Folding $\uparrow$ (\%)} & \textbf{MRT Easy $\uparrow$ (\%)} & \textbf{MRT Hard $\uparrow$ (\%)} & \textbf{Navigation $\uparrow$ (\%)} & \textbf{Orientation $\uparrow$ (\%)} & \textbf{Spatial Relations $\uparrow$ (\%)} & \textbf{Overall $\uparrow$ (\%)} \\
\rowcolor{violet!25}
\textbf{Number of Instances} & \textbf{200} & \textbf{200} & \textbf{200} & \textbf{400} & \textbf{400} & \textbf{400} & \textbf{1800} \\
\midrule
\rowcolor{white} {SmolVLM (500M)} & 34.00 & 34.00 & 22.00 & 32.50 & 51.00 & 30.00 & 35.22 \\
\rowcolor{violet!10} MiniCPM-V-2.6 (8B) & 27.50 & \textbf{36.00} & 29.00 & 9.75 & 45.00 & 14.75 & 25.72 \\
\rowcolor{white} InternVL2.5 (8B) & 41.50 & 30.00 & 24.50 & 33.25 & 69.00 & 64.50 & 47.72 \\
\rowcolor{violet!10} Llava-1.5 (7B) & 38.00 & 32.50 & 20.00 & 24.75 & 52.50 & 50.25 & 38.39 \\
\rowcolor{white} Idefics3 (8B) & 42.00 & 31.00 & \textbf{29.50} & \textbf{35.50} & 63.00 & 58.50 & 46.28 \\
\rowcolor{violet!10} InternVL2.5 (26B) & 42.00 & 31.50 & 28.00 & 28.25 & 70.00 & \textbf{70.75} & \textbf{48.83} \\
\rowcolor{white} o1 (Undisclosed)& 36.50 & 33.00 & 20.50 & 33.25 & 71.00 & 64.75 & 47.56 \\
\rowcolor{violet!10} Qwen2.5VL (7B) & 37.50 & 27.00 & 25.00 & 16.50 & 66.00 & 55.75 & 40.67 \\
\rowcolor{white} Qwen2.5VL (3B) & \textbf{46.00} & 31.50 & 25.50 & 13.50 & 63.10 & 33.25 & 35.86 \\
\rowcolor{violet!10} SmolVLM (3B) & 35.50 & 32.00 & 29.00 & 29.75 & 51.75 & 38.50 & 37.39 \\
\rowcolor{white} Llama-3.2-Vision (11B) & 26.00 & 29.50 & 24.00 & 26.25 & 48.00 & 48.75 & 36.17 \\
\rowcolor{violet!10} InstructBLIP (7B) & 36.50 & 32.00 & 21.00 & 10.50 & 51.00 & 17.50 & 27.50 \\
\rowcolor{white} InstructBLIP (13B) & 38.50 & 35.60 & 23.40 & 12.50 & 53.00 & 20.50 & 29.94 \\
\rowcolor{violet!10} GPT-4o (Undisclosed) & 36.00 & 32.50 & 20.00 & 32.75 & \textbf{72.50} & 66.50 & 47.44 \\
\rowcolor{white} LlavaNext (7B) & 27.00 & 32.50 & 26.50 & 27.00 & 52.00 & 52.50 & 38.78 \\
\rowcolor{violet!10} Random & 33.33 & 33.33 & 25.00 & 25.00 & 50.00 & 25.00 & 32.37 \\
\bottomrule
\end{tabularx}
% \vspace{-5pt}
\caption{\textbf{Accuracy of Models in Spatial Reasoning Tasks:} Presented are the accuracy percentages of different models across various spatial reasoning tasks, including Paper Folding, MRT Easy, MRT Hard, Navigation, Orientation, Relations, and Overall performance. The overall score is determined by taking a weighted average (based on the proportion of each subset) of these results, with the highest scores in each column emphasized in bold.}
% \vspace{-12pt}
\label{tab:performance}
\end{table*}

\section{Evaluation}

In our experiments, we use PyTorch~\cite{Torch} and Hugging Face Transformers~\cite{hf}. We evaluated the spatial intelligence of 13 prominent VLMs, which include both open-source and commercial variants. Among the commercial standards, we tested OpenAI's GPT-4o and o1~\cite{openai2024gpt4technicalreport, jaech2024openai}. The open-source models evaluated comprised Qwen2.5 3B and 7B~\cite{qwen2025qwen25technicalreport}, Llava 1.5 7B~\cite{Liu_2024_CVPR}, LlavaNext 7B~\cite{li2024llava}, InstructBlip 7B and 13B~\cite{dai2023instructblipgeneralpurposevisionlanguagemodels}, Idefics 8B~\cite{laurençon2024buildingbetterunderstandingvisionlanguage}, SmolVLM 500M and 2B~\cite{smolvlm}, Llama-3.2-Vision 11B~\cite{dubey2024llama}, MiniCPM-V-2.6 8B~\cite{yao2024minicpm}, and InternVL2\_5 8B and 26B~\cite{chen2024expanding}. All the model variants are instruction-tuned and greedy decoding~\cite{germann2003greedy} was equipped. To carry out these experiments, we use the Azure OpenAI API service for accessing OpenAI's models, and for the open-source models, we perform inference using 4$\times$ A100 40GB NVIDIA GPUs.

\subsection{Benchmark Distribution}
% As shown in Figure~\ref{fig:distribution}, 
The benchmark dataset comprises a total of 1,800 image-question pairs. Of these, 22.2\% (400 pairs) evaluate mental rotation ability, with equal distributions of 11.1\% (200 pairs) from the classic Mental Rotation Test (MRT) and 11.1\% (200 pairs) from the MRT easy set. For spatial visualization assessment, 11.1\% (200 pairs) feature stimuli with either single or double folds and between one and three pierced holes. The remaining pairs are evenly distributed, with 22.2\% (400 pairs) sampled from \emph{Maze-nav} to evaluate navigation skills, 22.2\% (400 pairs) from \emph{Spatial-Obj}~\cite{shiri-etal-2024-empirical} to assess spatial relations comprehension. Finally, 22.2\% (400 pairs) are equipped from \emph{EgoOrientBench}~\cite{jung2024right} to evaluate orientation understanding.

\subsection{Performance Evaluation}

\textbf{Overall Performance:}  
Table~\ref{tab:performance} summarizes the performance of various models across multiple spatial reasoning tasks, including two variants of the Mental Rotation Test (MRT), Paper Folding, Spatial Relations, Navigation, and Orientation. The chance level for these tasks is 32.37\%, which represents the expected accuracy if a model were to guess randomly, based on the number of choices available in each task. Notably, models such as InternVL2.5 (26B) and InternVL2.5 (8B) achieve the highest overall accuracy–48.83\% and 47.72\% respectively–surpassing this chance level by nearly 15\%, while several models (e.g., InstructBLIP variants and MiniCPM-V 2.6) perform near or significantly below chance levels overall, suggesting that their broader spatial reasoning capabilities remain limited.

\textbf{Spatial Relations:}  
In the Spatial Relations task, models demonstrate strong performance, with InternVL2.5 (26B) reaching 70.75\% and InternVL2.5 (8B) 64.50\%, indicating that they effectively capture relational aspects within scenes.

\textbf{Mental Rotation Tests (MRT):}  
The MRT tasks reveal disparities. In the easier version MRT, MiniCPM-V-2.6 (8B) leads with 36.00\%, yet its overall spatial reasoning performance is low (25.72\%). In the hard MRT variant, Idefics3 (8B) scores 29.50\%, with most models clustering between 20.00\% and 29.00\%. This suggests that success in a simplified mental rotation task does not necessarily translate to robust spatial reasoning across tasks.

\textbf{Paper Folding:}  
The Paper Folding task yields modest scores (with Qwen2.5VL (3B) at 46\% and both Idefics and InternVL2.5 (26B) around 42\%), reinforcing the notion that many models struggle with visualizing object transformations.

\textbf{Navigation and Orientation:}  
Navigation scores, while variable, are measured against an expected chance level of 50\% (because the tasks typically present two-choice questions). While certain models exceed this baseline, others—like Idefics3 (8B), achieving merely 35.50\%—perform worse than random guessing. In the Orientation task, models such as GPT-4o and o1 achieve scores above 70\%. This indicates that when visual cues like clear object outlines are present, some models can effectively discern orientation.

% \paragraph{Summary:}  
% Overall, these varied outcomes highlight that, although aspects of spatial reasoning (e.g., spatial relations and orientation) are captured by current models, more complex tasks such as mental rotation and visualization continue to pose significant challenges.

\textbf{This raises a critical question:}
\textit{Is the limitation in mental rotation due to the synthetic nature of the images lacking essential visual cues, or does it reflect a fundamental inability of VLMs to reason about spatial transformations?}

\begin{figure}[th!]
    \centering
    \includegraphics[width=\columnwidth]{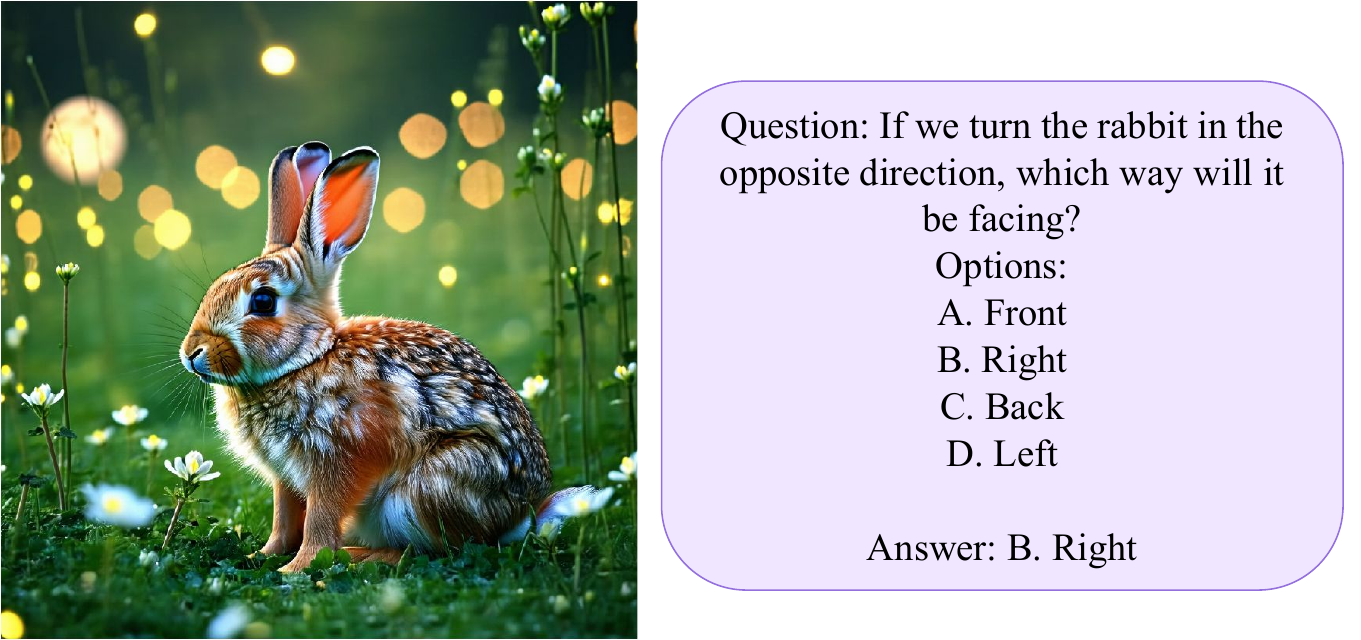}
    \vspace{-10pt}
    \caption{Example of a GenAI image shows a rabbit in a green meadow, challenging mental rotation skills. It asks VLMs to identify the rabbit's orientation after rotation, with "Right" as the correct choice.}
    \label{fig:genai}
    \vspace{-10pt}
\end{figure}

To explore this question, we examined a question-answer test centered around the use of generative AI (genAI)-produced images to evaluate mental rotation further. We developed a synthetic dataset employing generative models to examine mental rotation capabilities under controlled conditions (see Appendix~\ref{sec:app_img}). Initially, we crafted five meticulously designed image descriptions portraying objects in various orientations (e.g., facing right). These descriptions served as in-context learning prompts to Claude3.7~\cite{claude}, accompanied by a list of objects, to stimulate the generation of 100 analogous image descriptions using the specified objects or their synonyms. These generated descriptions were then processed by two generative models: \texttt{Flux.1-dev}~\cite{flux2024} and \texttt{Stable Diffusion 3.5 Large}~\cite{stabilityai2024sd35}, producing a total of 200 images. Following manual curation, we chose 80 question-answer pairs that necessitated inferring an object's orientation after a hypothetical rotation (refer to Figure~\ref{fig:genai} for illustrative examples).

\begin{figure*}[h!]
    \centering
    \includegraphics[width=0.9\linewidth]{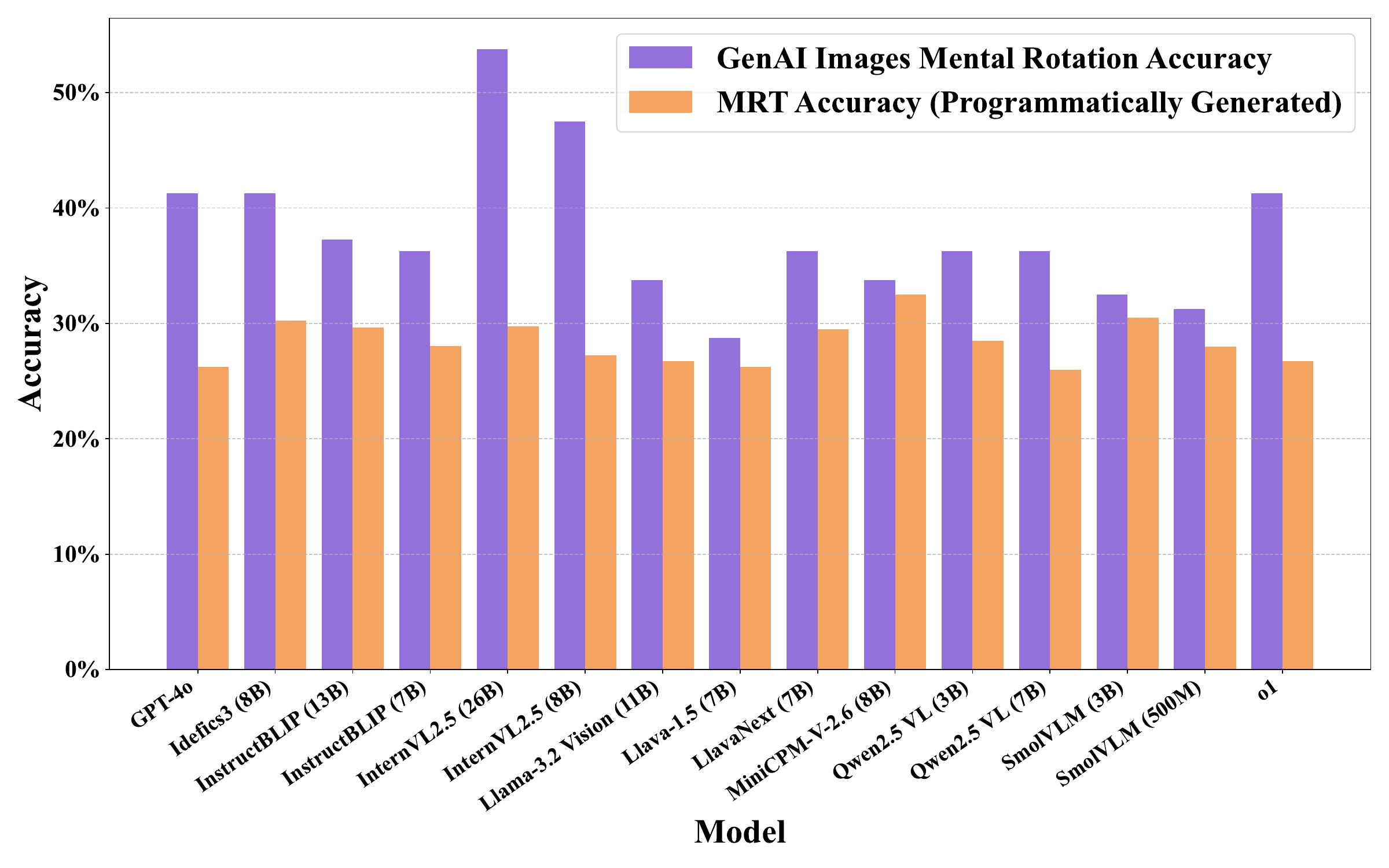}
    \vspace{-12pt}
    \caption{Comparison of Mental Rotation Images Accuracy (violet) and MRT Accuracy (teal) across various vision-language models. Models such as \textbf{InternVL2.5 (26B)} exhibit a significant gap between the two accuracy metrics, suggesting a stronger capability in interpreting mental rotation images compared to structured MRT tasks. In contrast, lower-performing models like \textbf{LLaVa-1.5 (7B)} show consistently weak performance across both categories. The overall trend indicates that while some models excel in image-based spatial reasoning, their understanding of abstract mental rotation tasks remains limited.}
    \vspace{-10pt}
    
    \label{fig:ablation}
\end{figure*}

As can be observed in Figure~\ref{fig:ablation}, performance on the synthetic mental rotation assessment varied considerably across models. Some models, particularly InternVL2.5 (26B) and InternVL2.5 (8B), consistently demonstrated higher overall accuracy compared to other models such as Llava-1.5 (7B) and SmolVLM (500M). Models like Llama-3-2 Vision (11B) and MiniCPM-V-2.6 (8B) exhibited relatively consistent performance between the GenAI-generated mental rotation task and the standard MRT. This suggests that performance differences are not solely due to the data type used. Choosing synthetic images similar to training data does not significantly improve performance, indicating a limitation in spatial reasoning. %The observed improvement with GenAI-synthetic images may be due to reliance on spurious correlations, not true spatial understanding.
The enhancement observed in the GenAI-synthetic images may be attributed to their potential resemblance to the original training dataset of these models or reliance on other visual cues.

\textbf{Summary:} While tasks such as Spatial Relations and Orientation reveal certain strengths of current models, the struggles in Mental Rotation, Paper Folding, and Navigation tasks highlight significant gaps in spatial reasoning. The novel genAI mental rotation assessment provides additional insights into these limitations and offers a controlled environment to further probe the mechanisms behind spatial transformations in VLMs.

%% file: sec/5_related_work.tex
\section{Related work}

\subsection{Spatial Reasoning in Vision-Language Models}
% Spatial understanding has been a rising topic addressed by various works, with most of them emphasizing on endowing VLMs with spatial understanding with added information using point clouds, scene graphs, and depth maps~\cite{chen2024spatialvlm, cheng2025spatialrgpt}. 

Recent progress in VLMs has significantly advanced multimodal comprehension, though explicit spatial reasoning still presents substantial challenges. Initially, VLMs were primarily tailored for overarching image understanding and captioning, frequently overlooking the intricate spatial relationships required for uses in robotics and augmented reality. As a countermeasure, various strategies have emerged, integrating spatial supervision into training datasets and model frameworks.

One area of research endeavors to create large-scale synthetic spatial reasoning datasets. For instance, \cite{chen2024spatialvlm} employs an automated 3D spatial VQA data production process to craft millions of region-conscious question–answer pairs derived from 2D imagery by building 3D scene graphs and incorporating metric depth estimation. This method enhances the training data with spatial annotations, thereby equipping the VLMs with improved spatial reasoning capabilities both qualitatively and quantitatively. In a similar vein, \cite{cheng2025spatialrgpt} expands on this concept by embedding region-level hints and relative depth data into the visual encoder. By integrating a depth-to-language connector and processing user-defined region proposals, the approach shows significantly improved outcomes on spatial reasoning benchmarks—even in intricate 3D environments. This design emphasizes the necessity of combining local region features with depth indicators to capture both broad contexts and precise spatial details.

Recent initiatives such as \cite{tang2024sparkle} focus on training VLMs in core 2D spatial tasks, enhancing skills such as direction interpretation, distance estimation, and localization, thus improving spatial reasoning. This approach suggests basic spatial skills lay the foundation for tackling complex challenges. Research into grounded and compositional strategies, such as multimodal spatial grounding, further improves alignment between visuals and language \cite{rajabi2023grounded}. However, models still fall short of human-level reasoning, especially in dynamic environments, indicating a need for future exploration. Improving VLM spatial reasoning requires effective data curation and architecture. Despite progress through techniques like 3D annotations and depth features, achieving reliable human-level understanding in real-world applications needs further effort.

\subsection{Spatial Reasoning in Humans}
Spatial reasoning is a multifaceted cognitive ability that enables individuals to perceive, manipulate, and navigate space. Seminal work by \citet{Shepard1971-np} introduced the mental rotation paradigm, laying the groundwork for subsequent studies that have refined our understanding of spatial cognition. Researchers such as \citet{hegarty2004dissociation} and \citet{newcombe2000making} have differentiated between intrinsic skills (e.g., mental rotation and spatial visualization) and extrinsic skills (e.g., navigation and perspective-taking), establishing frameworks that underscore the link between early spatial abilities and later academic achievement in STEM domains \citep{wai2009spatial}.

Recent intervention studies demonstrate that targeted spatial training can enhance children’s mathematical performance \citep{uttal2013malleability, cheng2014spatial}. In parallel, interdisciplinary research has applied computational and qualitative frameworks to model human spatial reasoning for applications in areas such as human–robot interaction and geographic information systems \citep{moratz2006spatial, montello1993scale}. These combined efforts affirm that spatial reasoning is not only a trainable and critical cognitive skill but also a pivotal foundation for solving real-world problems and advancing STEM education.

\subsection{Spatial Reasoning Benchmarking}
Benchmarking spatial reasoning capabilities is critical for evaluating the effectiveness of VLMs in real-world scenarios. Recent efforts have introduced dedicated benchmarks that focus on both qualitative and quantitative aspects of spatial understanding. For example, \cite{cheng2025spatialrgpt} not only improves model performance but also introduces a benchmark dataset comprising both qualitative and quantitative spatial reasoning tasks derived from indoor, outdoor, and simulated environments. This benchmark evaluates models on tasks such as determining relative positions (e.g., above, below, left, right) and measuring metric distances (e.g., direct, horizontal, vertical distances).

Other benchmarking approaches, such as those incorporated in \cite{tang2024sparkle}, focus on isolating basic spatial capabilities (direction, distance, localization) and then composing these to solve more complex spatial problems. Meanwhile, grounded spatial reasoning evaluations in multi-modal settings assess a model’s ability to align visual evidence with textual spatial descriptions \cite{rajabi2023grounded}. These benchmarks not only serve to highlight the current limitations of VLMs but also provide clear metrics for tracking progress as new architectures and training strategies are developed.

Concurrent work~\cite{xu2025defining} used human-applied psychometric tests to investigate spatial thinking in VLMs, with similar results. 
%Their results show that VLMs perform worse than humans on these tests, highlighting the need for more research into these models' spatial capabilities. Our method is different mainly because we use real-world photographs in addition to the psychometric assessments. 
%
Their results demonstrate that VLMs underperform relative to humans on these tests, underscoring the need for further exploration of these models' spatial capabilities. Our approach diverges by incorporating real-world photographs alongside psychometric assessments. To ensure full control over the images and expand the test cases, we also introduce custom-developed psychometric assessments.
%In order to have total control over the images and provide a greater number of test cases, we additionally develop unique psychometric exams. 
%
Finally, we perform an ablation study on the image data to see if any important cues that are necessary for the model to perform these tasks are absent.

Collectively, these benchmarking efforts underscore the need for systematic evaluation of spatial reasoning. They provide a foundation for comparing diverse approaches and guiding future research toward achieving robust, human-level spatial understanding in VLMs.

%% file: sec/6_conclusion.tex
\section{Conclusion}

This paper tackles the underexplored topic of spatial reasoning in VLMs by offering a clear definition, introducing a robust benchmark with both synthetic and real-world images, and evaluating 13 state-of-the-art VLMs. The key finding is stark: the majority of contemporary VLMs perform near random chance on spatial reasoning tasks on our benchmark; but some appear to perform better on tasks that use natural or generative images that may also contain other visual cues. Our work hence exposes a major limitation in VLMs' ability to achieve human-like visual understanding. This gap has significant implications, as spatial reasoning is vital for practical applications. Our work made a positive step in this direction; it didn't explore in depth which cues models use in natural images which can be seen as a limitation. Looking ahead, future research should continue to explore the interaction between spatial reasoning components and their respective roles in complex spatial tasks, as well as the reliance on other visual cues to reasoning, paving the way for more sophisticated and human-like spatial reasoning capabilities in artificial intelligence systems.

%% file: sec/X_suppl.tex
\clearpage
\setcounter{page}{1}
\maketitlesupplementary

% \section{Rationale}
% \label{sec:rationale}
% % 
% Having the supplementary compiled together with the main paper means that:
% % 
% \begin{itemize}
% \item The supplementary can back-reference sections of the main paper, for example, we can refer to \cref{sec:intro};
% \item The main paper can forward reference sub-sections within the supplementary explicitly (e.g. referring to a particular experiment); 
% \item When submitted to arXiv, the supplementary will already included at the end of the paper.
% \end{itemize}
% % 
% To split the supplementary pages from the main paper, you can use \href{https://support.apple.com/en-ca/guide/preview/prvw11793/mac#:~:text=Delete%20a%20page%20from%20a,or%20choose%20Edit%20%3E%20Delete).}{Preview (on macOS)}, \href{https://www.adobe.com/acrobat/how-to/delete-pages-from-pdf.html#:~:text=Choose%20%E2%80%9CTools%E2%80%9D%20%3E%20%E2%80%9COrganize,or%20pages%20from%20the%20file.}{Adobe Acrobat} (on all OSs), as well as \href{https://superuser.com/questions/517986/is-it-possible-to-delete-some-pages-of-a-pdf-document}{command line tools}.

\begin{table*}[!t]
\centering
\renewcommand{\arraystretch}{1.2} % Increase row spacing
\begin{tabular}{>{\raggedright\arraybackslash}p{0.9\textwidth}}
\toprule
\textbf{Examples} \\
\midrule
\midrule
A photo-realistic image of a car, viewed from the front, with the wheels turned to the left, parked in a driveway, with a clear blue sky in the background. \\[0.5em]
\hline
A photo-realistic image of a mug, viewed from the side, with the handle on the right, filled with steaming coffee, on a wooden table, with a window in the background showing a sunny day. \\[0.5em]
\hline
A photo-realistic image of a laptop, viewed from the back, with the screen open. \\[0.5em]
\hline
A photo-realistic image of a bicycle, viewed from the side, with the front wheel turned to the right, parked on a cobblestone street, with a row of colorful houses in the background. \\[0.5em]
\hline
A photo-realistic image of a cat, viewed from the front, with the tail curled to the right, sitting on a windowsill, with a potted plant in the background. \\
\bottomrule
\end{tabular}
\caption{Photo-realistic Image Examples that were given to Claude 3.7 to generated similar ones.}
\label{table:examples}
\end{table*}

\section{Data Creation}
\label{sec:app_data}

In our study, we developed Python scripts to generate image sets tailored for cognitive evaluations involving paper folding and mental rotation tests. The code and obtained images will be made publicly available upon acceptance.

For the folded paper images, we employed the Python Imaging Library (PIL) to simulate the visual process of \textbf{paper folding}. Initially, we draw a white paper measuring $100 \times 100$ pixels onto a $120 \times 120$ pixel canvas. Using predefined drawing routines and fold reflection rules, our script recursively applies vertical, horizontal, or diagonal fold lines, generating intermediate views of each fold stage through polygon clipping operations (Algorithm~\ref{alg:folded_paper}). Holes are randomly punched into the final folded polygon. The script then calculates their unfolded positions via reflections, effectively doubling layers with each fold. Finally, we produce candidate images, comprising the correctly unfolded paper and two distractors—these distractors either omit a hole, mirror positions, or slightly rotate them. All generated views and candidates are combined into a single composite image, clearly separating folding stages in the top row and randomized candidate options in the bottom row. The corresponding metadata for each image is recorded and stored in a JSONL file. To ensure clarity, our approach restricts folds to half of the paper at a time and avoids mixing diagonal folds with vertical or horizontal ones.

In parallel, we crafted a script for mental rotation tests (\textbf{MRT}) involving complex polycube shapes, utilizing matplotlib's 3D plotting capabilities (Algorithm~\ref{alg:mental_rotation}). Our method begins by selecting polycube configurations ranging from basic shapes to intricate forms such as snake-like arrangements. For each configuration, the script generates a set of candidate images through transformations including rotations, mirrored reflections, or substitutions with visually similar shapes. Candidates consist of one correct rotation and two distractors—one involving mirrored rotations and the other substituting with a similar but incorrect shape. The final image arranges the original shape prominently at the top, with the candidates randomized below, deliberately concealing transformation details to assess mental rotation capabilities. Two variations are implemented: a challenging version mimicking traditional MRT setups in human cognition without visual aids and using rotations at angles of $60^\circ$, $90^\circ$, and $120^\circ$ around all axes, and a simplified version with visual cues and rotations restricted to $-90^\circ$, $90^\circ$, or $180^\circ$ along a single axis. Images and metadata are stored in JSONL files for structured evaluation.

\begin{algorithm}[H]
\caption{Generate Folded Paper Test Image}
\begin{algorithmic}[1]
    \State \textbf{Initialize:} Define drawing routines and fold reflection rules.
    \State \textbf{Recursive Folding:} 
         \begin{itemize}
             \item Start with an initial paper polygon.
             \item For each fold, clip the polygon (using a fixed midpoint) and save the intermediate view.
         \end{itemize}
    \State Obtain the final folded polygon.
    \State Generate random holes within the final polygon.
    \State Compute unfolded hole positions via reflection (doubling layers per fold).
    \State Create candidate images:
         \begin{itemize}
             \item \textbf{Correct:} Use unfolded holes.
             \item \textbf{Wrong:} Modify holes (e.g., remove one, mirror, or rotate).
         \end{itemize}
    \State Assemble a composite image with:
         \begin{itemize}
             \item Top row: Fold views.
             \item Bottom row: Candidate options (labeled A, B, C).
         \end{itemize}
    \State Save the composite image and record metadata.
\end{algorithmic}
\label{alg:folded_paper}
\end{algorithm}

\begin{algorithm}[H]
\caption{Generate Mental Rotation Test Images}
\begin{algorithmic}[1]
    \State \textbf{Initialize:} Define \texttt{SHAPES} and \texttt{SIMILAR\_MAPPING}.
    \For{each image to generate}
        \State Randomly select a shape and compute its vertices.
        \State Generate three candidate transformations:
            \begin{itemize}
                \item \textbf{Correct:} Rotate the shape.
                \item \textbf{Wrong 1:} Mirror then rotate.
                \item \textbf{Wrong 2:} Use a visually similar shape then rotate.
            \end{itemize}
        \State Shuffle candidate order and record the correct option.
        \State Plot the original shape and candidates in a grid.
        \State Save the image and append metadata.
    \EndFor
\end{algorithmic}
\label{alg:mental_rotation}
\end{algorithm}

\begin{table*}[!t]
\centering
\begin{minipage}{0.9\textwidth}
\raggedright
\textbf{Role:} Text-Based Image Description Generation Assistant \\[1ex]
\textbf{Objective:} To generate high-quality text image descriptions for generative models. \\[1ex]
\textbf{Input (Textual):}
\begin{itemize}[leftmargin=1.5cm]
    \item A list of objects (provided as text by the user).
    \item Example image descriptions (provided as text by the user).
\end{itemize}
\textbf{Output (Textual):} New image descriptions (as text). \\[1ex]
\textbf{Task:} Generate new text image descriptions that meet the following criteria:
\begin{enumerate}[leftmargin=1.5cm]
    \item \textbf{Style Mimicry:} Replicate the writing style, sentence structure, and vocabulary used in the example text descriptions.
    \item \textbf{Object Novelty:} Feature the provided list of objects, ensuring they are different from objects explicitly mentioned in the example text descriptions.
    \item \textbf{Setting Novelty:} Describe the objects in new and different settings or contexts compared to those presented in the example text descriptions.
    \item \textbf{Logical Coherence \& Realism:} Ensure all generated text descriptions are logically sound, realistic, and portray plausible scenarios in text. Avoid nonsensical or physically impossible descriptions in text.
\end{enumerate}
\end{minipage}
\caption{Image Description Generation Guidelines}
\label{tab:guidelines}
\end{table*}

\section{GenAI Image Generation Process}
\label{sec:app_img}

\begin{figure*}[t!]
    \centering
    \includegraphics[width=1\linewidth]{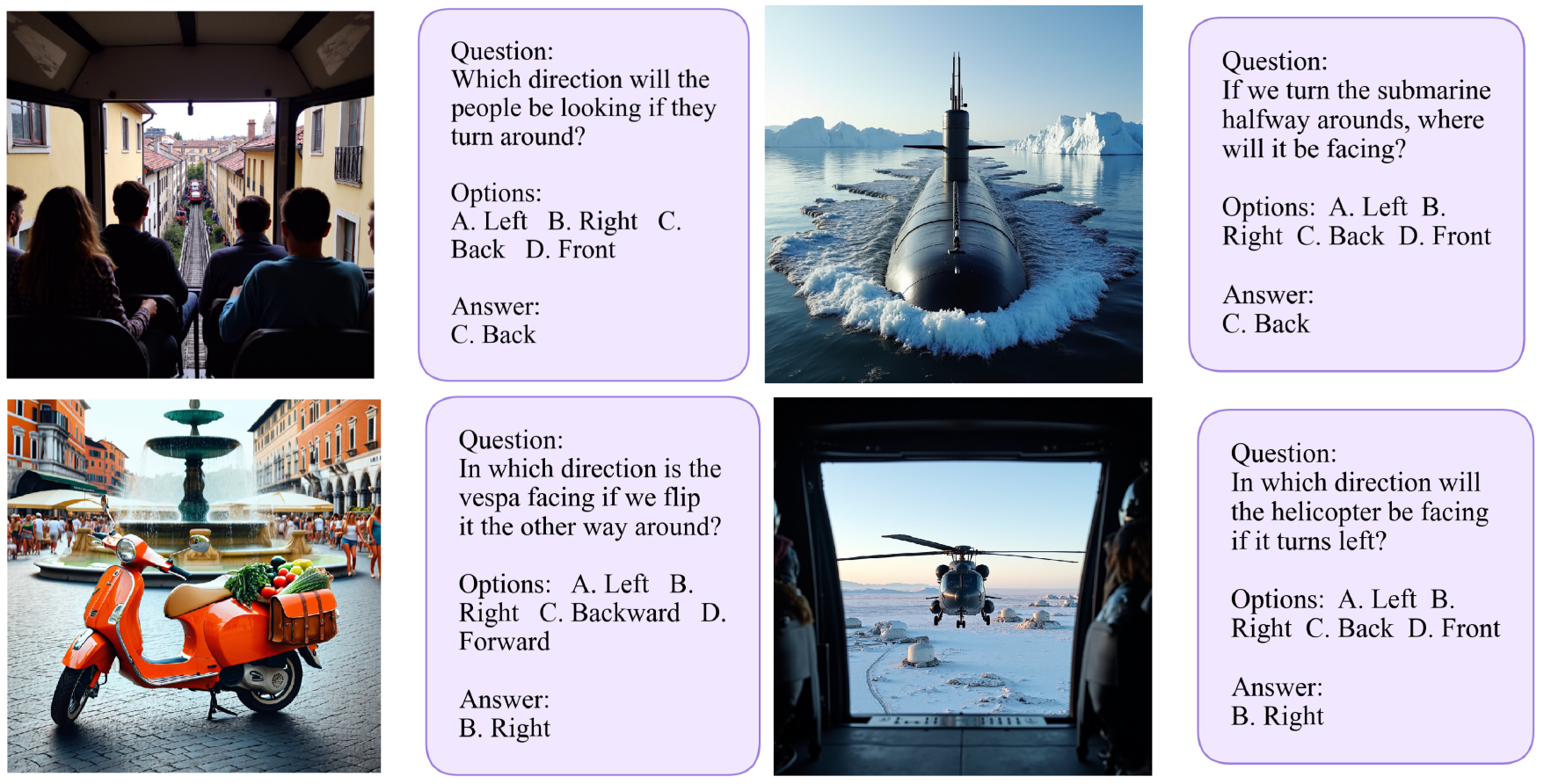}
    \caption{Example of GenAI-generated Images and Question-Answer pairs that were used for the mental rotation evaluation.}
    \label{fig:examples}
\end{figure*}

To create the images for our ablation study, we manually handcrafted five exemplar image descriptions, each depicting an object positioned in a specific orientation within a photo-realistic setting (see Table~\ref{table:examples}). We then fed these examples into \texttt{Claude3.7}, a Large Language Model (LLM) by Anthropic, to generate additional descriptions featuring different objects. The model was prompted using the structured format shown in Table~\ref{tab:guidelines}. This prompt is meticulously organized, beginning with clearly defined components such as the \textbf{Role} and \textbf{Objective}, which immediately establish the context and purpose of the task. It subsequently delineates the required \textbf{Input} and expected \textbf{Output} through bullet points, ensuring clarity and ease of understanding. The core of the prompt—the \textbf{Task}—is presented as a numbered list that outlines specific criteria, including style mimicry, object and setting novelty, and logical coherence and realism. This detailed segmentation not only clarifies the structure for the generated text but also systematically guides the generative process. Overall, the structured use of bold headings and lists enhances the prompt's effectiveness and user-friendliness by ensuring that all essential information is clearly highlighted and easily referenced.
Alongside the five exemplars, we provide a compiled list of objects to give extra context and guidance to the model. The list is the following: \texttt{bicycle}, \texttt{motorcycle}, \texttt{scooter}, \texttt{car} \texttt{truck}, \texttt{bus}, \texttt{train}, \texttt{airplane}, \texttt{helicopter}, \texttt{boat}, \texttt{ship}, \texttt{dog}, \texttt{cat}, \texttt{bird}, \texttt{fish}, \texttt{rabbit}. Figure~\ref{fig:examples} depicts examples of the generated images from the diffusion models alongside the hand-curated questions and answers.

\section{Implementation Details}
\label{sec:app_impl}

For the evaluation of these 13 state-of-the-art (SOTA) models, we utilized PyTorch~\cite{Torch} (version 2.5.1) along with the HuggingFace\cite{hf} libraries: \texttt{transformers}, \texttt{diffusers}, and \texttt{accelerate}. Each model was loaded using \texttt{bfloat16} precision, and \texttt{Flash Attention 2}~\cite{dao2023flashattention2fasterattentionbetter} was implemented wherever applicable to enhance inference speed. The experiments were executed on four A100 Nvidia GPUs, each with 40 GB of memory, using CUDA version 12.0. Models were used with a greedy decoding approach, providing their outputs immediately, and not utilizing any prompting methods such as Chain-of-Thought (CoT)~\cite{wei2022chain} or Least-To-Most (LTM)~\cite{zhou2022least} prompting. The results were matched using regular expressions, and responses that did not conform to the specified format were deemed incorrect.